# A Temporal Logic For Uncertain Events And An Outline Of A Possible Implementation In An Extension Of PROLOG

*Soumitra Dutta*

Computer Science Division
University of California
Berkeley, CA 94720

dutta@ernie.berkeley.edu

## ABSTRACT

There is uncertainty associated with the occurrence of many events in real life. In this paper we develop a temporal logic to deal with such uncertain events and outline a possible implementation in an extension of PROLOG. Events are represented as fuzzy sets with the membership function giving the possibility of occurrence of the event in a given interval of time. The developed temporal logic is simple but powerful. It can determine effectively the various temporal relations between uncertain events or their combinations. PROLOG provides a uniform substrate on which to effectively implement such a temporal logic for uncertain events.

## 1 Introduction

The concept of time plays an important role in our life and thus it is natural that temporal knowledge is required in a wide range of disciplines including computer science, business, engineering, psychology, philosophy and linguistics. In recent years there has been a greater emphasis among researchers in computer science ( especially from the fields of artificial intelligence and databases) on the development of adequate models for representing temporal knowledge. Most of the developed models address the issue of representing temporal knowledge in a world devoid of *uncertainity*, i.e., we are sure of the occurrence or non-occurrence of a particular event at some instant of time. Such a classically *crisp* approach is useful and sometimes sufficient, but often lead to awkward or sometimes even wrong solutions when applied to many real world situations, most of which are inherently uncertain. Consider a simple example:

*John went to the market to do his shopping. On returning home, he found that his key was missing.*

Here, the event of John losing his key has occurred, but we are not sure whether John lost his key on the way to the market or while buying groceries or on his way back. Of course another possibility is that he lost his key even before he left for the market or after returning home ( e.g., accidentally dropped it on the doorstep). Several other events can be uncertain in this simple example, e.g., John may have also bought some groceries at a small shop on his way to (or back from ) the market besides at the market. Now we might desire to determine temporal

This research has been supported in part by NASA Grant No. NCC-2-275 and NSF Grant NSF DCR 8513139.

relations between events, some of which are uncertain, e.g., how possible is it that John lost his key before going to the supermarket. Also, we would like to be able to determine temporal relationships between *combinations* of complex events, e.g., how possible is that John lost his key while *going to the market or shopping at the market* where the combining operator *or* is the usual non-exclusive or. Note that due to the uncertainity in the occurrence of the events, it is only natural that the model be able to specify the desired temporal relationships by the degree to which they are possible.

Such kind of situations are suspiciously familiar in the real world and call for a temporal model which can incorporate this inherent uncertainity while making inferences. There are two requirements which any such model should satisfy: first, it should provide an adequate representation for uncertain events and second, it should allow the development of a suitable calculus for determining temporal relationships between uncertain events. In this paper we develop a model for representing and manipulating temporal knowledge incorporating uncertainity and outline a possible implementation of these ideas in an extension of PROLOG. We shall be more concerned about uncertainty in the occurrence of events as opposed to the uncertainty in the duration of events. We shall use *fuzzy logic* [11, 12, 13] for representing uncertainity in our temporal model. An earlier paper [4] describes the theoretical ideas presented in this paper in greater detail. Due to limitations on the length of this paper, we present here only the bare essential details, skipping an overview of past research and examples. Also omitted are the various axioms and theorems of the proposed temporal logic which are described in detail with examples in [4].

The layout of this paper is as follows. Section two describes the fundamentals of our proposed model. Section three develops a simple but powerful calculus for manipulating events in the developed model. Section four outlines a possible approach to implementing our proposed temporal logic model in an extension of PROLOG. Finally, section five provides a brief conclusion.

## 2 Definitions Of Basic Temporal Relationship Operators

In this section, we shall briefly present the fundamentals of our proposed temporal model.

Let I represent the set of time intervals. Small letters i,j,k,.. or $i_1, i_2,..$ refer to individual intervals.

$$I = \{i_1, i_2, \cdots, i_n\}$$

For simplicity we assume that the intervals i, $i \varepsilon I$ are disjoint in time. We place no other restriction on the construction of the time intervals.

E represents the universe of events. Small letters e,f,g,.. or $e_1, e_2,..$ refer to individual events.

$$E = \{e_1, e_2, \cdots, e_m\}$$

Events are represented as fuzzy sets over the universe I, the membership function giving the possibility of occurrence of the event in any given interval. The *possiblity* that an event e occurs in interval i is given by $\mu_i(e)$. Note that $0 \leq \mu_i(e) \leq 1$. If $\mu_i(e) = 0$, then event e cannot occur in interval i. Similarly if $\mu_i(e) = 1$, then event e definitely occurs in interval i.

**Definition 1** : The occurrence of an event is defined by the *possibility* of its occurrence in any interval i, $i \varepsilon I$. Formally,

$$e = \bigcup_{i \varepsilon I} \{<i, \mu_i(e)>\}$$

The support set of an event e consists of all intervals i, $i \varepsilon I$, in which it is possible that event e occurs.

**Definition 2** : The support set $S_e$ of an event e is defined as

91

$$S_e = \bigcup_{i \in I, \mu_i(e) > 0} \{<i>\}$$

Note that $(\forall\ i \in I - S_e)(\mu_i(e) = 0)$ Figure 1 illustrates the classical *crisp* interpretations of the various temporal operators. Note that as we are considering disjoint sets of intervals, other temporal relationships are possible, but we only consider the six common temporal relationships [9] here.

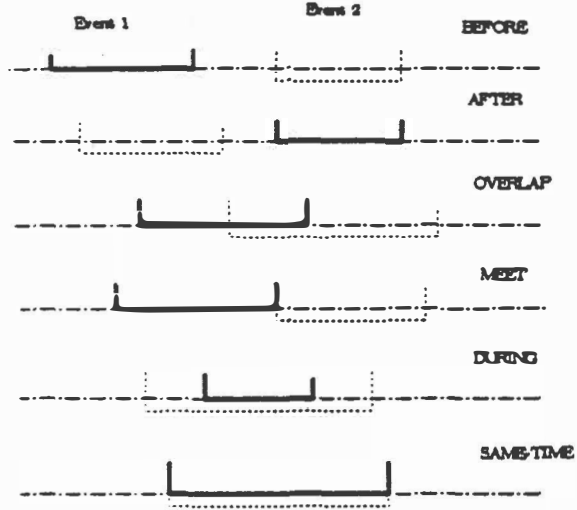

**Figure 1** : The Various Different Temporal Relationships

Corresponding to these six different temporal relationships, we can define six temporal operators. Let us first consider the temporal relations *before, after, overlap* and *meet*. < can be considered as a *precedence* operator and $\mu_<(e_1, e_2)$ gives the degree to which event $e_1$ precedes event $e_2$. > is the inverse of the < operator and $\mu_>(e_1, e_2)$ gives the degree to which the event $e_1$ follows event $e_2$. o is the *overlap* operator and $\mu_o(e_1, e_2)$ gives the degree to which event $e_1$ overlaps event $e_2$. m is the *meet* operator and $\mu_m(e_1, e_2)$ gives the degree to which event $e_1$ meets event $e_2$, i.e., the degree to which event $e_1$ immediately precedes event $e_2$.

For giving a formal definition of the above fuzzy temporal operators, we need to define certain relationships on intervals.

**Definition 3** : For i, $i \in I$ let $i_-$ and $i_+$ represent the time of the beginning and end of interval i.

$\xi_p(i_1, i_2) = 1$ iff $i_1$ ends before $i_2$ begins.

$\xi_\omega(i_1, i_2) = 1$ iff there does not exist another distinct time interval between $i_1$ and $i_2$.

Now we can define the required fuzzy temporal operators.

**Definition 4a** :

$$\mu_<(e_1, e_2) = \max_{i_j \in S_{e_1},\ i_k \in S_{e_2},\ \xi_p(i_j, i_k) = 1}(\min(\mu_{i_j}(e_1), \mu_{i_k}(e_2)))$$

$$\mu_>(e_1, e_2) = \max_{i_j \in S_{e_1},\ i_k \in S_{e_2},\ \xi_p(i_k, i_j) = 1}(\min(\mu_{i_j}(e_1), \mu_{i_k}(e_2)))$$

$$\mu_o(e_1, e_2) = \max_{i \in S_{e_1} \cap S_{e_2}}(\min(\mu_i(e_1), \mu_i(e_2)))$$

$$\mu_m(e_1, e_2) = \max_{i_j \in S_{e_1},\ i_k \in S_{e_2},\ \xi_\omega(i_j, i_k) = 1}(\min(\mu_{i_j}(e_1), \mu_{i_k}(e_2)))$$

Now we provide the definitions for temporal relationships *same-time* and *during*.

**Definition 4b:** Two events $e_1$ and $e_2$ occur at the *same-time* if they are temporally equivalent, i.e.,

$$\bigcup_{i \in S_{e_1}} \{<i, \mu_i(e_1)>\} = \bigcup_{i \in S_{e_2}} \{<i, \mu_i(e_2)>\}$$

**Definition 4c:** Event $e_1$ occurs during $e_2$ if the following condition is satisfied:

$$(\forall \ i \in S_{e_1})(\mu_i(e_1) \leq \mu_i(e_2))$$

These definitions for the various temporal relationship operators are based on an intuitive extension of the definitions for the classical crisp case where we are sure of the occurrence or non-occurrence of an event at any given time. It can be verified that these definitions provide the desired answers in the crisp case. It is possible to provide alternative definitions for these operators which may be more meaningful in certain situations.

## 3 Combining Events

In this section, we present the definitions for combining events to form *compound events*. Note that events are closed under the operations of complementation, intersection and union described below. Thus we can use the temporal relationship operators described in the last section to determine approximate temporal relationships between *compound events*.

**Complement of an Event**

The complement of a event $e_c$ is defined as

**Definition 5 :**

$$e_c = \bigcup_{i \in I} \left\{ <i, 1-\mu_i(e)> \right\}$$

**Intersection of Events**

The intersection of $e_1$ and $e_2$ is defined iff $S_{e_1} \cap S_{e_2} \neq \Phi$. If $S_{e_1} \cap S_{e_2} = \Phi$ then $e_1 \cap e_2$ is not defined.

**Definition 6 :**

$$S_{e_1} \cap S_{e_2} \neq \Phi \rightarrow e_1 \cap e_2 = \bigcup_{i \in S_{e_1} \cap S_{e_2}} \left\{ <i, \min(\mu_i(e_1), \mu_i(e_2))> \right\}$$

**Union of Events**

**Definition 7 :**

$$e_1 \cup e_2 = e = \bigcup_{i \in S_{e_1} \cup S_{e_2}} \left\{ <i, \max(\mu_i(e_1), \mu_i(e_2))> \right\}$$

**Example 1**

Let us present a brief example to illustrate some of the ideas presented till now. Consider the simple statements:

*John wanted to buy some groceries. He went to the market. On returning home he found that he had lost the key to his house.*

93

Let us define (arbitrarily) three intervals of time,

$i_1$ = time for going to the market
$i_2$ = time spent in the market
$i_3$ = time spent coming back home

and four events

$e_1$ = travelling away from home
$e_2$ = buying goods
$e_3$ = travelling towards home
$e_4$ = losing key to house

Let us represent the degree of occurrences of the various events during the various intervals in the form of a matrix with the $a_{ij}$ element giving the possibility of occurrence of the $i^{th}$ event during the $j^{th}$ interval.

|       | $i_1$ | $i_2$ | $i_3$ |
|-------|-------|-------|-------|
| $e_1$ | 0.8   | 0.2   | 0     |
| $e_2$ | 0.2   | 0.6   | 0.2   |
| $e_3$ | 0     | 0.2   | 0.8   |
| $e_4$ | 0.5   | 0.7   | 0.5   |

We can compute some of the temporal relationships between events by using definition 4. For example, the degree of possibility of John losing the key to the house before buying goods is given by

$$\mu_<(e_4,e_2) = \max(\min(0.5,0.6),\min(0.5,0.2),\min(0.7,0.2)) = 0.5$$

Similarly, the degree of possibility of John losing the key to the house before leaving home for the market is given by

$$\mu_<(e_4,e_1) = \max(\min(0.2, 0, 0)) = 0.2$$

Suppose we wish to know the possibility that John lost the key while traveling:

Travelling = $e = e_1 \cup e_3$
Desired answer = $\mu_o(e_4,e) = 0.5$

Suppose we wish to know the possibility that John lost the key either before he left home or after he returned home:

event of before leaving home and after returning home = $f = (e_1 \cup e_2 \cup e_3)_c$.
Desired answer = $\mu_o(e_4,f) = 0.4$

The important difference between our model and the classical *crisp* models proposed by Allen [3] and other researchers is that in our model uncertain events are represented as fuzzy sets and the operators {<,>,o,m} need not take on only binary values (true/false). This greatly enhances the the power of the conclusions drawn by the system, e.g., Allen's model cannot quantify the degree of overlap of two events. Scheng [9] and Kahn [5] also developed temporal models for handling uncertain events. Scheng's model was more powerful than the one developed by Kahn, but it does not provide techniques for combining events and determining temporal relationships between them.

## 4 Implementation Of The Proposed Temporal Logic In An Extension Of PROLOG

In this section, we outline a possible implementation of the proposed temporal logic in an extension of PROLOG. We are in the process of implementing these ideas and thus some of the final details may differ from those described below.

Predicates of PROLOG now correspond to the events described in the previous sections on temporal logic. In normal PROLOG, each predicate is assumed to be universally true, i.e., it is devoid of any temporal qualifications. While extending PROLOG to handle our proposed temporal logic we extend predicates to be possibly true to certain degrees in different intervals of time (see definition 1). Thus the general syntax of a PROLOG fact and rule become respectively :

$$\text{predicate\_A } \{<i_A, \mu_{i_A}(A)>\}$$

and

$$\text{predicate\_A } \{<i_A, \mu_{i_A}(A)>\} :\text{- predicate\_B } \{<i_B, \mu_{i_B}(B)>\}$$

We adopt similar notation as before. Let $S_A$ represent the support set of predicate_A, i.e., the set of intervals over which predicate_A is possibly true.

For implementing out proposed temporal logic, we have to add three features to ordinary PROLOG. First, we have to expand the syntax of PROLOG to handle the representation of temporal information. We assume that this can be taken care of relatively easily by some uniform *adhoc* notational scheme. For the purposes of this exposition, we shall informally refer to temporally qualified predicates by listing the intervals over which they are temporally qualified in a generic set form, e.g., predicate_A can be represented as

$$\text{predicate\_A } \{ <i_j, \mu_{i_j}(A)> \}$$

where $\mu_{i_j}(A)$ represents the degree to which it is possible that A is true in interval $i_j$. Second, we have to augment the unification procedure of the PROLOG interpreter to handle temporal information and provide a rule for propogating the uncertain temporal information from the antecedents to the consequents (i.e., from the RHS to the LHS of a standard PROLOG rule written in the form A:-B). Finally, we have to provide some meta-level predicates to implement the temporal relationship operators described in sections 3 and 4 and illustrated in figure 1. We shall describe the latter two additions now.

### Augmenting the Unification Procedure

For the unification of

$$\text{predicate\_A } \{ < i \text{ sub } A, \mu_{i_A}(A) > \} \text{ and}$$
$$\text{predicate\_A' } \{ < i \text{ sub } A', \mu_{i_{A'}}(A') > \}$$

to succeed two conditions must be satisfied:

First the normal unification of predicate_A and predicate_A' must succeed. This does not require any modification in existing PROLOG interpreters besides an ability to accept the syntax of temporally qualified predicates. Second, the temporal unification of predicate_A and predicate_A' must succeed. Generally this unification is approximate and shall succeed only to a degree $\tau$. Three cases arise:

i. $S_A \subset S_{A'}$ such that

$$(\forall \ i \varepsilon S_A)(\mu_i(A) \le \mu_i(A'))$$

In this case the temporal unification of predicate_A and predicate_A' succeeds to the degree 1, i.e., $\tau = 1$.

ii. It is the case that

$$(\exists i \varepsilon S_A)(\mu_i(A) \le \mu_i(A'))$$

In this case, the temporal unification of predicate_A and predicate_A' succeeds to a

95

degree $\tau$ where $\tau$ is defined as follows. Let S, $S \subset S_A$, represent the set of intervals for which the following condition is satisfied.

$$(\forall\ i \varepsilon S)(\mu_i(A) \le \mu_i(A'))$$

$D_S$ and $D_{S_A}$ represent the durations of S and $S_A$ respectively, where $D_S$ and $D_{S_A}$ are given by:

$$D_S = \sum_{i \varepsilon S} (i_+ - i_-) \times \mu_i(A)$$

and

$$D_{S_A} = \sum_{i \varepsilon S_A} (i_+ - i_-) \times \mu_i(A) \text{ and}$$

where $i_-$ and $i_+$ respectively represent the beginning and end of interval i. Then we can define $\tau$ as

$$\tau = \frac{D_S}{D_{S_A}}$$

iii. The third and final case is that either

$$S_A \cap S_{A'} = 0.$$

or

$$(\forall\ i \varepsilon S_A)(\mu_i(A) > \mu_i(A'))$$

In this case the degree of success of the temporal unification is 0, i.e., $\tau = 0$ and we can consider that the unification of the two predicates fails.

The implementation of this temporal unification can be achieved by generating a goal

temp_unify ( pred_A { $< i_A\ ,\ \mu_{i_A}(A) >$ } , pred_A' { $< i_A\ ,\ \mu_{i_{A'}}(A') >$ }, $\tau$)

where $\tau$ is the value of the degree of success achieved in the temporal unification of the two predicates and is computed according to three different possible conditions specified above. This meta-level goal for the interpreter is generated upon the success of the normal unification of predicate_A and predicate_A'. We also need a rule for propogating the degree of temporal unification from the antecedents to the consequent of a PROLOG rule. Various different formulations are possible for defining such a propogation rule. We choose the simple *min* rule of fuzzy set theory which has been used by other researchers [13] while extending PROLOG to handle uncertainty. Consider the PROLOG rule:

predicate_A { $< i_A\ ,\ \mu_{i_A}(A) >$ } :- $\Lambda_i$ ( $predicate\_A_i$ { $< i_A\ ,\ \mu_{i_{Ai}}(A_i) >$ } )

where $\Lambda_i$ represents normal conjunction of PROLOG goals. Let $\tau_{A_i}$ represent the degree of success of temporal unification of $predicate\_A_i$. Then $\tau_A$, the degree of sucess of temporal unification of predicate_A is given by

$$\tau_A = \min_i(\tau_{A_i})$$

If $\tau_A = 0$, then we can consider the goal predicate_A as having failed.

### Adding Meta Level Predicates

The temporal relationships illustrated in figure 1, can be implemented in PROLOG by providing suitable meta level predicates for computing the desired relationships. The generic form of these meta-level predicates shall be:

NAME_OF_META_LEVEL_PREDICATE ( INP_PRED_1, INP_PRED_2, $\tau$)

where $\tau$ gives the degree to which the two input predicates, INP_PRED_1 and INP_PRED_2 satisfy the specified meta_level predicate. For example

BEFORE ( predicate_A { $< i_A, \mu_{i_A}(A) >$ } , predicate_A' { $< i_A\ ,\ \mu_{i_{A'}}(A') >$ }, $\tau$)

computes the degree $\tau$ to which predicate_A temporally precedes predicate_A'. This

96

computation is performed according to definition 4 of section 2 and a piece of PROLOG code to perform this computation can be easily written. We shall need at least 6 such meta level temporal relationship predicates corresponding to the 6 different temporal relationships illustrated in figure 1. The definitions of these various temporal relationship operators were specified in section 2. All these definitions are simple and easily implemented by PROLOG code. These meta-level temporal relationship predicates can be used in normal rules in our extended PROLOG system, e.g.,

$$\text{pred\_A :- pred\_B , pred\_C , before( pred\_B, pred\_C, } \tau_{before} \text{ )}.$$

where for simplicity we have not indicated the intervals over which the predicates A, B and C are temporally qualified. Assuming the success of temporal unification of predicates B and C are given by $\tau_B$ and $\tau_C$ respectively, then the success of the head predicate A is given by

$$\tau_A = min(\tau_B, \tau_C, \tau_{before})$$

where $\tau_{before}$ gives the degree to which predicate B temporally precedes predicate C and is computed according to definition 4.

## 5 Summary

In this paper, we have presented the bare essentials of a simple but powerful temporal logic for dealing with uncertain events. Events have been represented as fuzzy sets with the membership function giving the possibility of the occurrence of the event in a time interval. An axiomitization of the developed temporal model and other relevant details can be found in [4]. We have also outlined a possible implementation of our temporal logic for uncertain events in an extension of PROLOG.